\documentclass[10pt,twocolumn,letterpaper]{article}

\usepackage{iccv}
\usepackage{times}
\usepackage{epsfig}
\usepackage{graphicx}
\usepackage{amsmath}
\usepackage{amssymb}
\usepackage{eso-pic}
\usepackage[ruled,linesnumbered]{algorithm2e}
\usepackage{algpseudocode}
\usepackage{caption}
\usepackage{booktabs}
\usepackage{wrapfig}
\usepackage{subfigure}
\usepackage[accsupp]{axessibility}  

\usepackage[pagebackref=true,breaklinks=true,letterpaper=true,colorlinks,bookmarks=false]{hyperref}

\iccvfinalcopy 

\def\iccvPaperID{6804} 

\ificcvfinal\fi

\begin{document}

\title{Online Refinement of Low-level Feature Based Activation Map for Weakly Supervised Object Localization}

\author{Jinheng Xie, Cheng Luo, Xiangping Zhu, Ziqi Jin, Weizeng Lu, Linlin Shen\thanks{Corresponding Author}\\
School of Computer Science \& Software Engineering, Shenzhen University, China\\
Shenzhen Institute of Artificial Intelligence of Robotics of Society, Shenzhen, China\\
Guangdong Key Laboratory of Intelligent Information Processing, Shenzhen University, China\\
{\tt\small xiejinheng2020@email.szu.edu.cn, xiangping.zhu2010@gmail.com, llshen@szu.edu.cn}
}

\maketitle
\ificcvfinal\thispagestyle{empty}\fi

\begin{abstract}
    We present a two-stage learning framework for weakly supervised object localization (WSOL). While most previous efforts rely on high-level feature based CAMs (Class Activation Maps), this paper proposes to localize objects using the low-level feature based activation maps. In the first stage, an activation map generator produces activation maps based on the low-level feature maps in the classifier, such that rich contextual object information is included in an online manner. In the second stage, we employ an evaluator to evaluate the activation maps predicted by the activation map generator. Based on this, we further propose a weighted entropy loss, an attentive erasing, and an area loss to drive the activation map generator to substantially reduce the uncertainty of activations between object and background, and explore less discriminative regions. Based on the low-level object information preserved in the first stage, the second stage model gradually generates a well-separated, complete, and compact activation map of object in the image, which can be easily thresholded for accurate localization. Extensive experiments on CUB-200-2011 and ImageNet-1K datasets show that our framework surpasses previous methods by a large margin, which sets a new state-of-the-art for WSOL. Code will be available soon.
\end{abstract}
\vspace{-6pt}
\section{Introduction}
\label{sec:intro}

Supervised object localization and detection methods based on deep neural networks~\cite{2017Faster,liu2016ssd,redmon2016you,lin2017feature} have achieved great advances. However, these methods usually rely on massive training data with intensive annotations, especially location-level labels. To alleviate the high annotation costs, weakly supervised object localization requiring only image-level annotations has gained lots of attentions. 

\begin{figure}[t]
	\begin{center}
		\includegraphics[width=\linewidth]{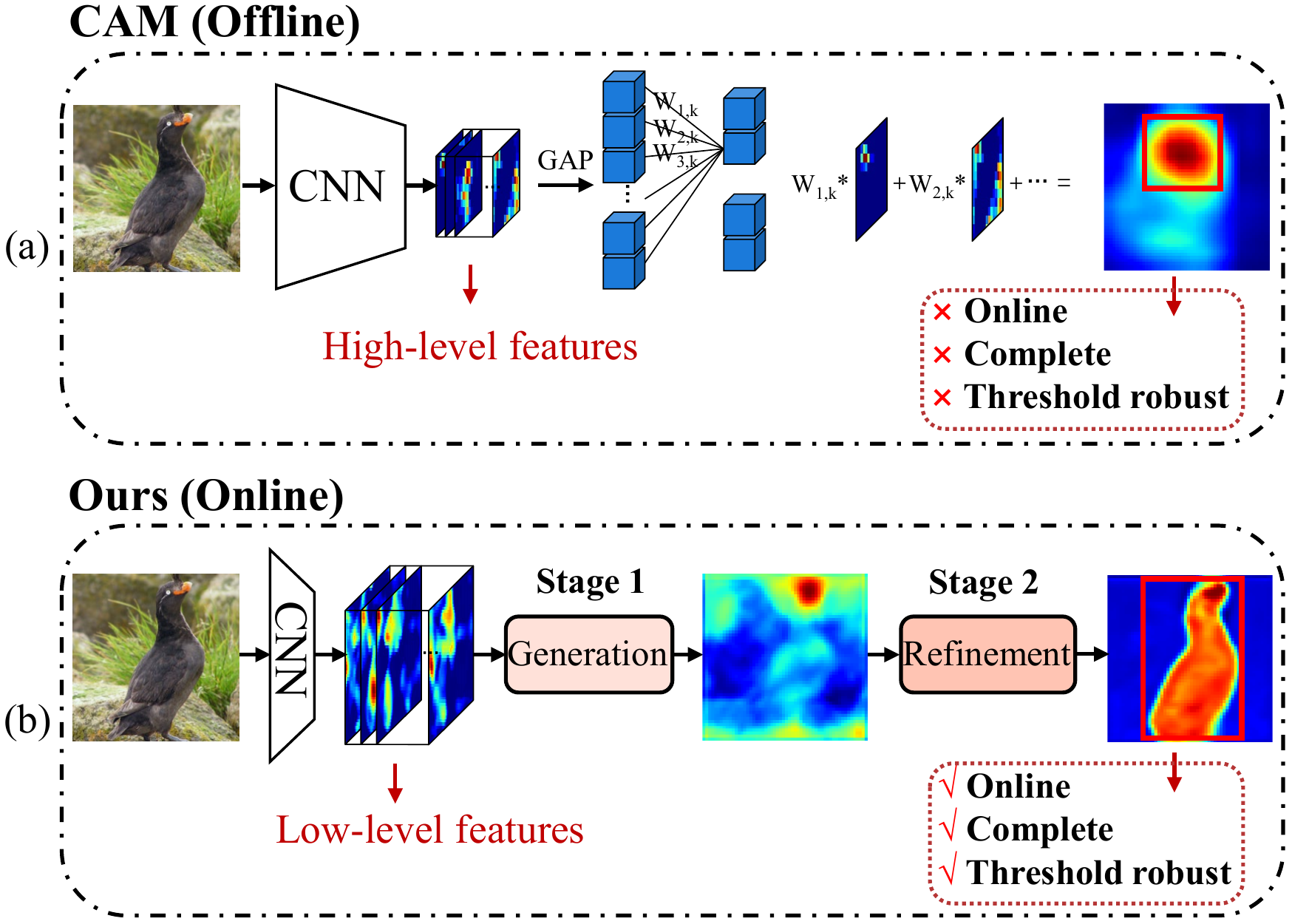}
	\end{center}
	\vspace{-21pt}
	\captionsetup{font=small}
	\caption{Comparison between CAM and the proposed method. (a) Overview of the CAM pipeline. (b) Overview of the proposed two-stage learning framework. Red bounding boxes illustrate the localization results.}
	\label{fig:idea_compar_wit_cam}
	\vspace{-18pt}
\end{figure}
\begin{figure}[t]
	\begin{center}
		\includegraphics[width=\linewidth]{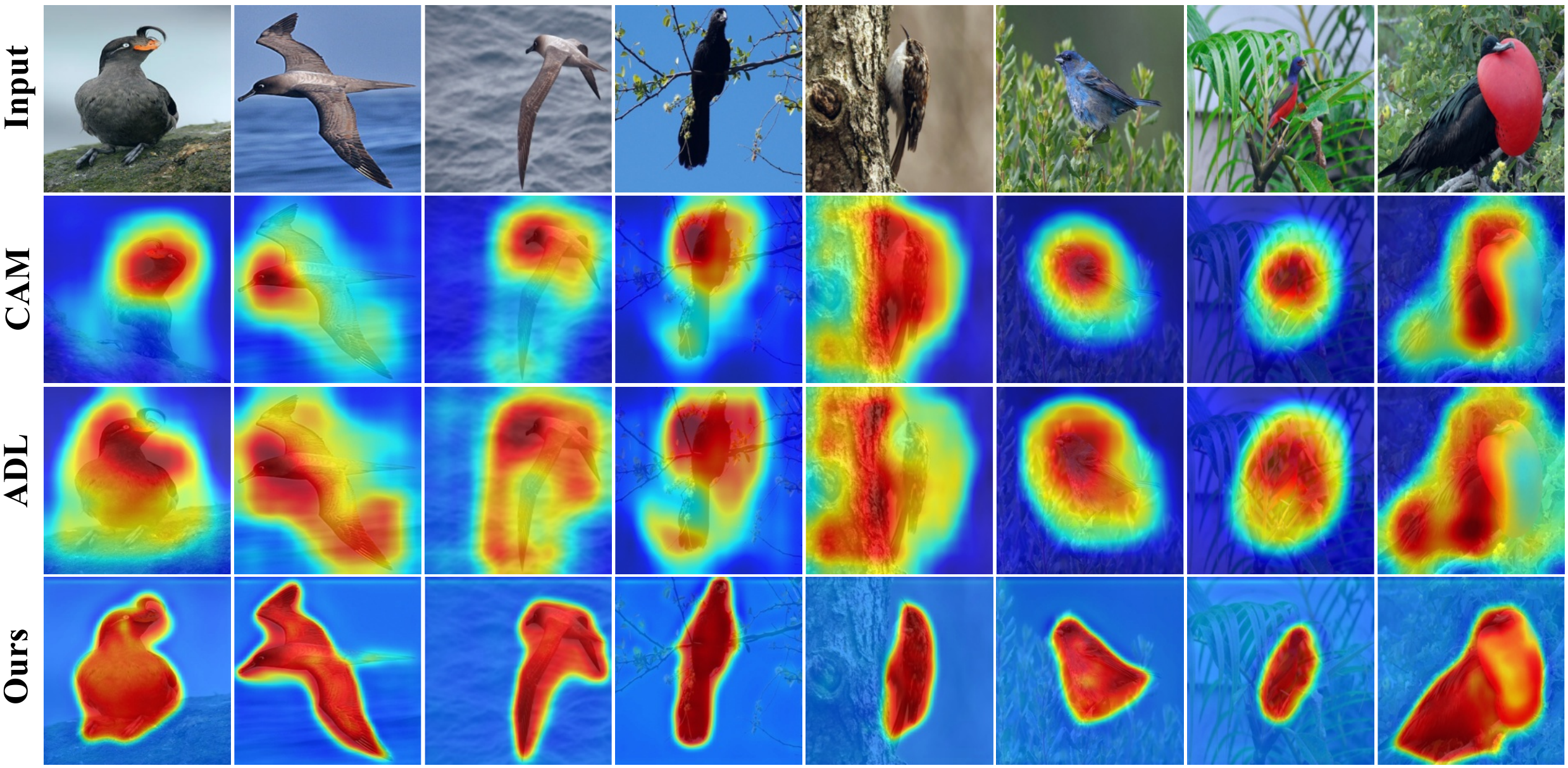}
	\end{center}
	\vspace{-21pt}
	\captionsetup{font=small}
	\caption{Visualization of activation maps of CAM, ADL, and ours. The images are from the CUB-200-2011 testing set.}
	\label{fig:cam_adl_ours}
	\vspace{-14pt}
\end{figure}

Instead of exploring the entire extent of the objects, classification networks incline to identify patterns from small and sparse regions. Due to this limitation, class activation maps~\cite{zhou2016learning} (CAM), a weighted average of high-level feature maps, usually indicate only the discriminative regions of objects. Fig.~\ref{fig:idea_compar_wit_cam}(a) gives the diagram of CAM. As shown in figure, only the head region, which is the most discriminative part to distinguish the bird, is localized. See Fig.~\ref{fig:cam_adl_ours} for more examples of CAM based localization. However, discriminative region is insufficient for accurate object localization. To address this issue, various solutions \cite{song2014on,cinbis2014multi,wang2014weakly,bazzani2016self,zhang2018self,singh2017hide,choe2019attention,zhang2018adversarial,zhang2020rethinking,GCNWSOJ} have been explored. For example, \cite{kim2017two,  zhang2018adversarial, choe2019attention} attempt to erase the most discriminative regions, promoting the network to discover less discriminative regions.
As high-level features are used as region guidance, these approaches have limited potentials to derive complete and compact activation maps.
Furthermore, when the following thresholding is used to separate foreground and background pixels, the result is very sensitive to the setting of threshold, due to the diverse distribution of activation maps across different objects and background.

The defects of CAM based solutions can be summarized as:
(1) The generation of CAM is offline and thus can not be easily refined online. (2) Abstract semantics of high-level features are difficult to produce both complete and compact activation maps.  (3) Due to the ambiguity between foreground and background pixels across different activation maps, the localization results are sensitive to the threshold used in post-processing.

\begin{figure}[t]
	\begin{center}
		\includegraphics[width=\linewidth]{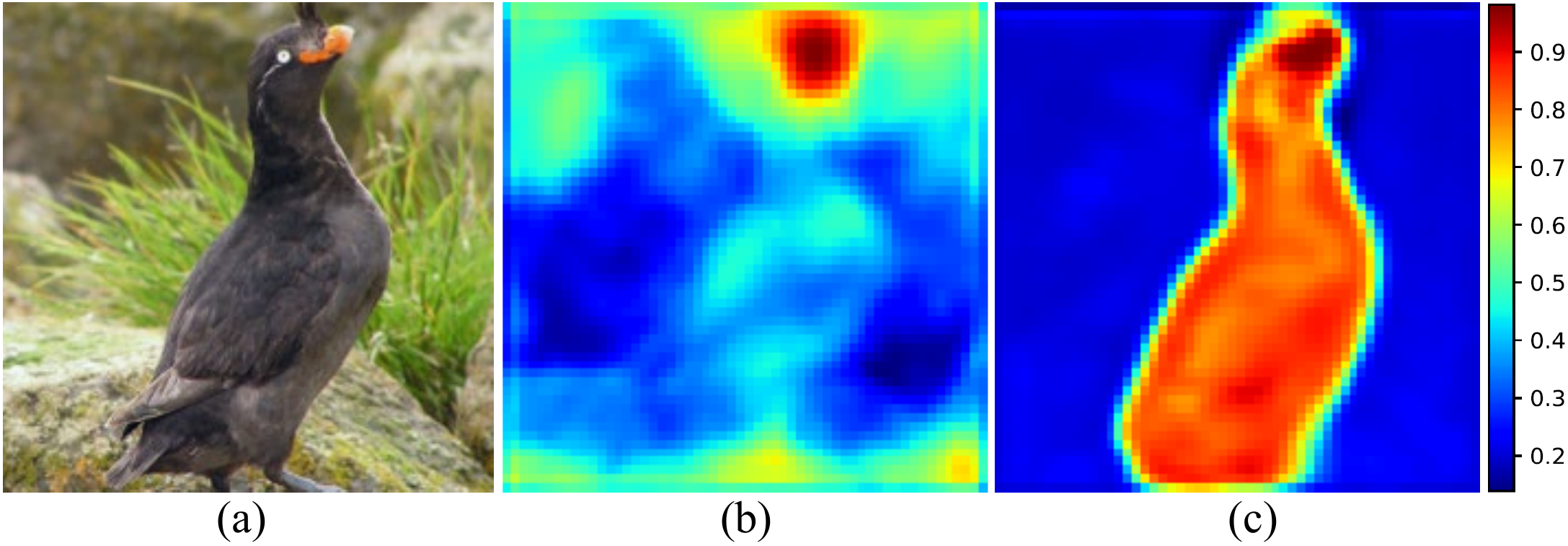}
	\end{center}
	\vspace{-21pt}
	\captionsetup{font=small}
	\caption{(a) An input image. (b) The coarse activation map from stage one. (c) The refined activation map from stage two. }
	\label{fig:cm_cmp}
	\vspace{-15pt}
\end{figure}

\begin{figure}[t]
	\begin{center}
		\includegraphics[width=\linewidth]{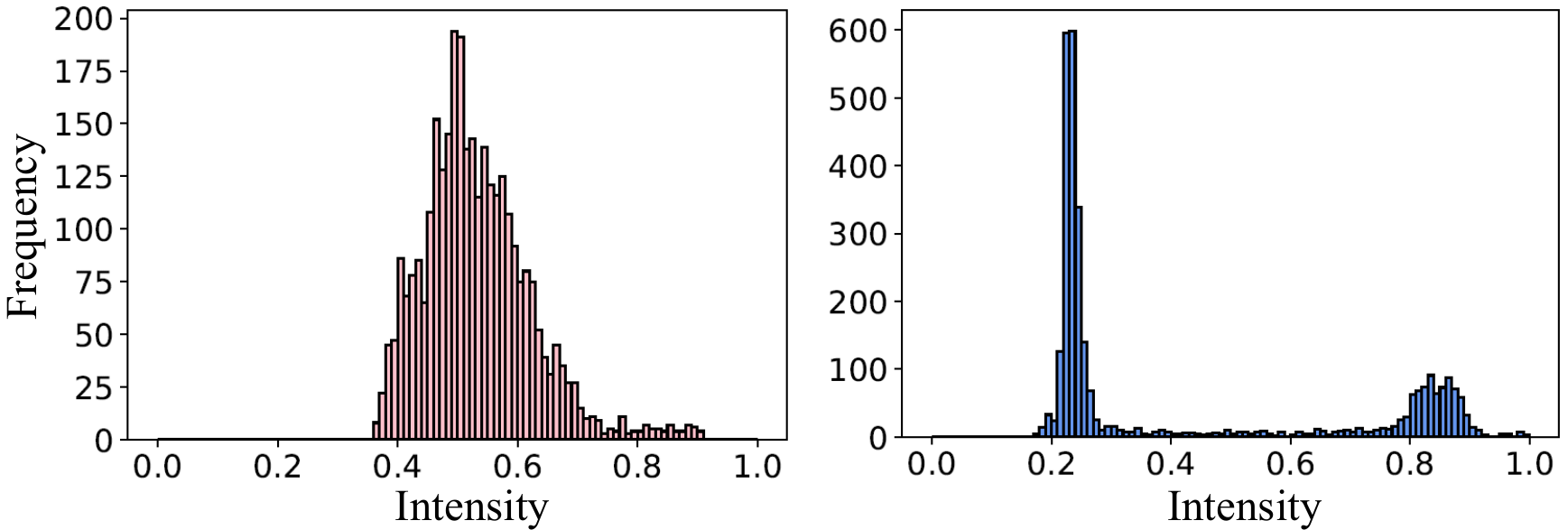}
	\end{center}
	\vspace{-21pt}
	\captionsetup{font=small}
	\caption{Histogram of activation maps. Left: the histogram calculated from Fig.~\ref{fig:cm_cmp}(b). Right: the histogram calculated from Fig.~\ref{fig:cm_cmp}(c). }
	\label{fig:hist_cmp}
	\vspace{-11pt}
\end{figure}

In this paper, we develop a two-stage learning framework for weakly supervised object localization, as shown in Fig.~\ref{fig:idea_compar_wit_cam}(b). Compared to high-level features, we argue that low-level features, which contain details or contour information of the object, are more suitable and effective region guidance for object localization. Based on this observation, we propose in the first stage low-level features based activation map generator, to explore those underlying information in the low-level features. It mainly contains an image classifier and an activation map generator (generator for short) with a classification head. The generator is directly incorporated into the shallow layer of the image classifier. Co-supervision of two classification losses leads to an online activation maps generation based on the low-level features. However, in most cases, these activation maps also suffer from two limitations. 1) The pixel value in activation maps intensively locates on around 0.5 (Fig.~\ref{fig:hist_cmp}(b)). When thresholding is applied to locate the object, localization results are sensitive to the threshold. 2) While more regions containing contour and context information are included in the low-level feature based activation map, discriminative regions like bird head still dominate the highly activated regions (Fig.~\ref{fig:cm_cmp}(b)). To address these issues, the second stage, called entropy-guided refinement, is proposed. The network architecture consists of a generator (initialized from the first stage) and an evaluator (a pre-trained image classifier). The evaluator aims to evaluate and ensure the quality of activation maps produced by the generator through an evaluation loss. Based on this, we further design a weighted entropy loss to alleviate the first limitation, by reducing the uncertainty of each single pixel in activation maps. Moreover, two components, including an attentive erasing and an area loss, are designed to deal with the second limitation. The two components can explore more alternative contents for classification and penalize the background pixels. Designs in the second stage adversarially encourage the generator to further explore the contextual semantics of objects preserved in the low-level features. Fig.~\ref{fig:cam_adl_ours} shows the comparison results of our method and two other existing popular object localization methods, $\ie$, CAM~\cite{zhou2016learning} and ADL~\cite{choe2019attention}. From Fig.~\ref{fig:cm_cmp}(b) to (c) and from left to right in Fig.~\ref{fig:hist_cmp}, it reveals that, after the entropy-guided refinement, our method can get well-separated, complete and compact object activation map, to achieve the accurate object localization.

Collectively, the main contributions of this paper can be summarized as:
\begin{itemize}
    \item We propose to employ low-level features as region guidance to generate activation maps in an online manner, which provides rich contextual information of objects for the following refinement.
    \item We design the entropy-guided refinement to adversarially drive the network to further explore the low-level features to obtain well-separated, complete, and compact activation maps for accurate object localization. As the activations of objects and background are now well-separated, the localization is not sensitive to the threshold.
    \item Extensive experiments on CUB-200-2011~\cite{wah2011caltech} and ImageNet-1K~\cite{russakovsky2015imagenet} datasets show that the proposed method surpasses the previous methods by a large margin, setting a new state-of-the-art performance. Experiments on additional datasets verify the robustness and generalization ability of our method across various scenarios and species.
\end{itemize}

\section{Related Work}
\textbf{Weakly Supervised Object Localization (WSOL)} aims to locate objects with less cost of annotation. Among various forms of weak supervision, the image-level label is mostly preferred by researchers. With only image-level labels, diverse solutions~\cite{song2014on,cinbis2014multi,wang2014weakly,bazzani2016self,zhou2016learning,zhang2018self,singh2017hide,choe2019attention,zhang2018adversarial,zhang2020rethinking,GCNWSOJ} have been explored to train deep neural networks for object localization.

\textbf{CAM based methods.}  \cite{zhou2016learning,singh2017hide,zhang2018adversarial,xue2019danet,choe2019attention} usually employ discriminative regions as guidance to locate the target objects. These class activation maps, a weighted average of high-level feature maps, approximate spatial distribution of discriminative regions. Unfortunately, the active responses in class activation maps only cover the most discriminative regions, instead of the entire object. The random erase of image patches \cite{singh2017hide} helps to locate less discriminative regions of objects. Furthermore, \cite{kim2017two,li2019guided,choe2019attention, meil} develop the erasing technique to drop the most discriminative regions.

Other than the above solutions, self-produced guidance (SPG)~\cite{zhang2018self} progressively generates object masks in a stage-wise manner. It employs high confident object regions as the seeds of foreground and the supervision of the lower layers. By integrating foreground regions, it enforces classification networks to learn pixel correlations from multiple layers. However, uncertain regions are less explored. 

\textbf{Geometric constraint.} ~\cite{GCNWSOJ} designs a novel network architecture called GC-Net, which contains a detector, a generator, and a classifier. The detector predicts a set of location coefficients, which are transformed by the generator to a 2D mask. Then the input image is split into foreground and background regions using the mask. During training, the categorical cross-entropy is adopted to minimize the uncertainty of object classification and the negative entropy loss is used to maximize the uncertainty of background classification. In this way, the mask gradually approaches the target object. However, the training of GC-Net might be unstable, due to the abstract transformation from numerical values to the 2D matrix. Besides, this approach has less potential to be applied to multi-instance localization.

\textbf{Attention mechanism.} Deep networks with attention mechanisms focus on informative semantics. Since the proposal of attention~\cite{vaswani2017attention,wang2018non,ParkWLK18,Woo_2018_ECCV,hu2018squeeze,self-at}, various tasks (\eg classification and detection) have involved attention mechanism for better features learning.

\cite{choe2019attention} utilizes an attention mechanism to hide the most discriminative parts and stochastically highlight the informative regions. However, a hard drop of the most discriminative part loses pixel correlations and expanding cues. In contrast, in our setting, we add constraints to the areas to be dropped and drop pixels with adaptive probabilities.

Residual Attention Networks (RAN)~\cite{RAN} stacks multiple attention modules to capture mixed attention. 
Attention Branch Network (ABN)~\cite{Fukui_2019_CVPR} introduces a branch structure with an attention mechanism. This branch structure built on a top layer directly generates an attention map in the training iteration.  Likewise, our method designs a sub-network to produce the activation map simultaneously, but on the shallow layer.

\section{Methodology}
\begin{figure*}[t]
\vspace{-4pt}
\begin{center}
   \includegraphics[width=\linewidth]{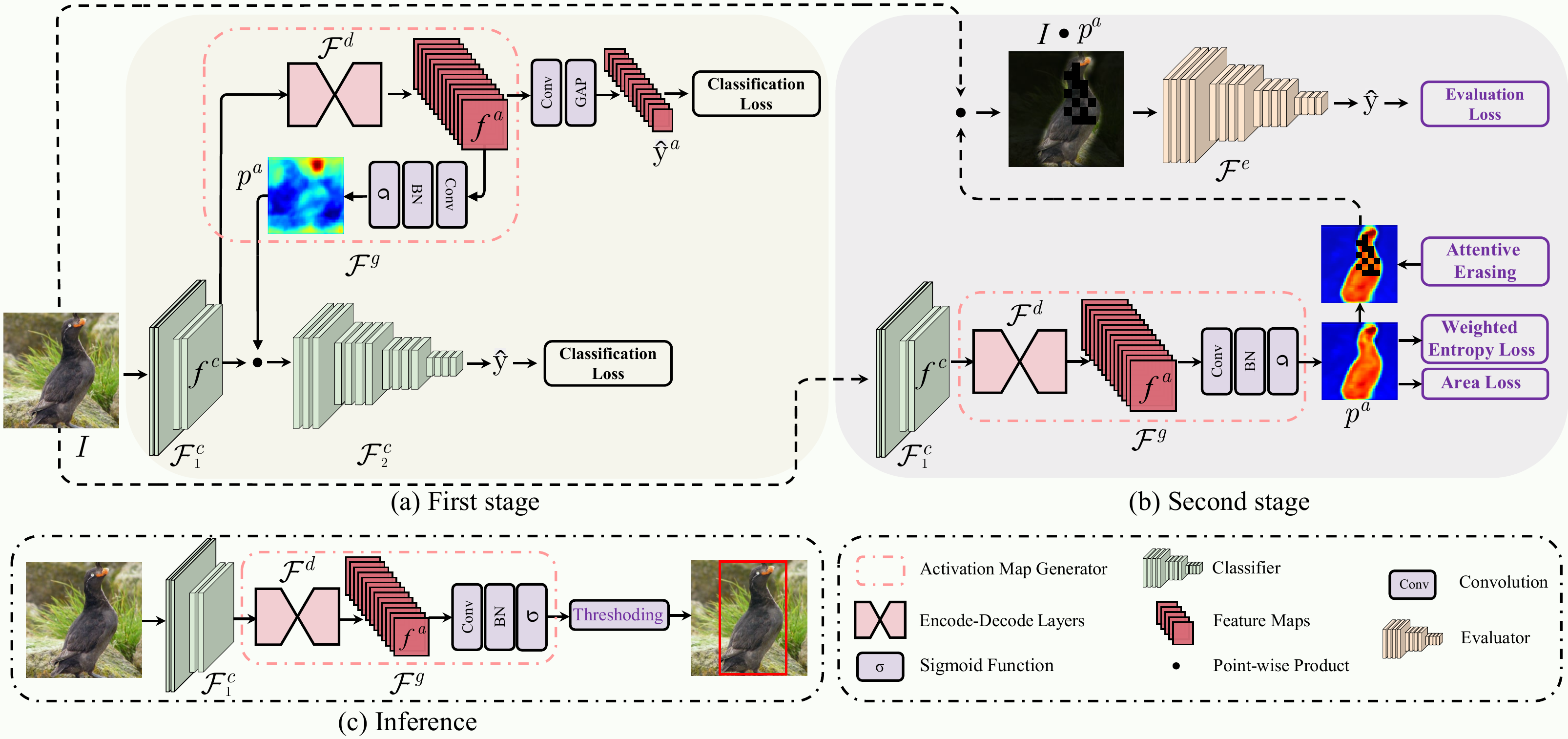}
\end{center}
    \vspace{-21pt}
   \captionsetup{font=small}
   \caption{Overview of the proposed two-stage learning framework. (a) First stage. Supervised by two classification losses, the generator $\mathcal{F}^g$, incorporated in the shallow layer of the classifier $\mathcal{F}^c$, turns the low-level features into coarse activation maps $p^a$. (b) Second stage. The network consists of a generator $\mathcal{F}^g$ (initialized from the first stage) and an evaluator $\mathcal{F}^e$. The evaluator aims to evaluate activation maps predicted by the generator through the evaluation loss. In addition, the weighted entropy loss, attentive erasing, and area loss are proposed to adversarially drive the generator to refine the coarse activation maps. (c) The inference for locating objects.}
   \vspace{-11pt}
\label{fig:network}
\end{figure*}

Details of the proposed two-stage learning framework are presented in Fig.~\ref{fig:network}. \textbf{First stage}: As shown in Fig.~\ref{fig:network}(a), the activation map generator with the classification head is integrated into the shallow layer of an image classifier (\eg ~\cite{resnet,VGG}). During training, supervised by two classification losses, the generator employs the low-level features to online yield activation maps of objects with rich contextual information. \textbf{Second stage}: After the first stage, we employ an evaluator (a pre-trained image classification network, \eg~\cite{VGG, resnet}), coupled with the generator (initialized from the first stage), to evaluate the generated activation maps (Fig.~\ref{fig:network}(b)). During training, three loss functions and an attentive erasing are proposed to supervise the model. In particular, the generator gradually yields well-separated, complete, and compact activation maps based on the contextual object information preserved in the low-level features. During inference, as shown in Fig.~\ref{fig:network}(c), only shallow layers of the classifier and the generator are needed to predict activation maps for object localization. In the following, we provide more details about the proposed framework.
\subsection{Low-level Feature Based Activation Map}
\textbf{First stage.} The designed activation map generator $\mathcal{F}^{g}$ is incorporated into shallow layer of the image classifier $\mathcal{F}^{c}$, which separates the classifier into two sub-networks $\mathcal{F}^{c}_1$ and $\mathcal{F}^{c}_2$ (as shown in Fig.~\ref{fig:network}(a)). Specifically, given an input image $I$, the low-level features can be derived as follow:
\begin{equation}
    f^c = \mathcal{F}^{c}_1(I\mathbf{; W}_{1}^c),
\end{equation}
 where $f^c \in \mathbb{R}^{h \times w \times c}$ is the low-level features. $\mathbf{W}^{c}_1$ represents the learning parameters of $\mathcal{F}^{c}_1$. $h,w,c$ denote the height, width, and number of channel of $f^c$, respectively. To generate the activation map $p^a$, $f^c$ is fed to the generator, which consists of encoder-decoder layers $\mathcal{F}^d$, a 2D convolution layer ($\mathbf{Conv}$), and a Batch Normalization layer ($\mathbf{BN}$):
 \begin{equation}
 f^a = \mathcal{F}^{d}(f^c;  \mathbf{W}^{d}), \quad
 p^a = \mathbf{BN(Conv}(f^a;\mathbf{W}^{p}_1)), 
\end{equation}
where $\mathbf{W}^{d}$ and $\mathbf{W}^p_1$ represent the learning parameters of $\mathcal{F}^{d}$ and $
\mathbf{Conv}$-$\mathbf{BN}$ layer, respectively, $f^a \in \mathbb{R}^{h\times w\times k}$ is the output features of $\mathcal{F}^{d}$ with $k$ channels, where $k$ denotes the number of classes. $p^a \in \mathbb{R}^{h \times w \times 1}$ is the generated activation map. Following the sub-network $\mathcal{F}^c_1$,  $\mathcal{F}^c_2$ aims to produce the class probability distribution $\hat{y}$:
\begin{equation}
     \hat{y} = \mathcal{F}^{c}_2(p^a\boldsymbol{\cdot} f^c; \mathbf{W}^c_{2}),
\end{equation}
where $\mathbf{W}^{c}_2$ represents the learning parameters of $\mathcal{F}^{c}_2$,  $p^a\boldsymbol{\cdot} f^c$ is the point-wise product between $p^a$ and $f^c$. $p^a$ act as the mask to mask out the background clutters in $f^c$. 
To get more accurate activation maps, an auxiliary classification head, including one Global Average Pooling ($\mathbf{GAP}$) layer and one Convolution ($\mathbf{Conv}$) layer, is added following $\mathcal{F}^{d}$:
\begin{equation}
     \hat{y}^a = \mathbf{GAP(Conv}(f^a; \mathbf{W}^{p}_2)),
\end{equation}
in which $\mathbf{W}^{p}_2$ represents the learning parameters of $\mathbf{Conv}$.

With corresponding image-level one-hot encoding label $y$, the classification losses $\mathcal{L}_{c}(y, \hat{y})$ and $\mathcal{L}_{c}(y, \hat{y}^a)$, respectively corresponding to $\hat{y}$ and $\hat{y}^a$, are formulated as:
\begin{equation}
\label{eq:5}
   \mathcal{L}_{c}(y, \hat{y}) = -\sum_{i}^{k} y_{i} \log \left(\frac{e^{\hat{y}_{i}}}{\sum_{j}^{k} e^{\hat{y}_{j}}}\right),
\end{equation}
\begin{equation}
    \mathcal{L}_{c}(y, \hat{y}^a) = -\sum_{i}^{k} y_{i} \log \left(\frac{e^{\hat{y}_{i}^a}}{\sum_{j}^{k} e^{\hat{y}_{j}^a}}\right).
\end{equation}

$\mathcal{F}^{g}$ and $\mathcal{F}^{c}_2$ can work complementarily during network training. As aforementioned, before feeding $f^c$ into the classification subnetwork $\mathcal{F}^{c}_2$, $p^a$ is applied to mask out the background clutters in $f^c$. Thus, with the supervision of the label $y$, the generator $\mathcal{F}^{g}$ learns to explore the contextual semantics in the low-level features, \ie excite the object regions in $p^a$.

In addition, the auxiliary classification head following $\mathcal{F}^{g}$ can further encourage $\mathcal{F}^{g}$ to learn more underlying information of the object in the low-level features, such that activation map $p^a$ includes more details and contextual information.

The Attention Branch Network (ABN)~\cite{Fukui_2019_CVPR} has a similar architecture as our model. However, there are several differences: (1) We employ the low-level features as region guidance to generate the coarse activation map $p^a$. Compared with the high-level features used in ABN, the low-level features, persevering more details and contour information of the object, are more suitable for object localization. (2) The low-level features usually contain high-frequency noise, \eg, twigs, and gravels. Based on this observation, the encoder-decoder layers $\mathcal{F}^{d}$ is added to alleviate this problem. (3) Our network architecture is designed to preserve more information from low-level features. 



\subsection{Entropy-guided Refinement}
\textbf{Second stage.} As shown in Fig.~\ref{fig:network}(b), the network architecture consists of the shallow layers of the classifier, a generator, and an evaluator. Both the shallow layers and generator are initialized from the first stage. Supervised by the evaluation loss, the evaluator, following the generator, aims to evaluate the quality of the generated activation maps. Based on this, we further propose the weighted entropy loss, attentive erasing, and area loss to address the limitations aforementioned above.

Forward inference of the network can be formulated as:
\begin{equation}
    f^c = \mathcal{F}^{c}_1(I\mathbf{; W}_{1}^c),
\end{equation}
\begin{equation}
p^a = \mathcal{F}^{g}(f^c; \mathbf{W}^{p}),
\end{equation}
\begin{equation}
\hat{y} = \mathcal{F}^{e}(I\boldsymbol{\cdot} p^a; \mathbf{W}^{e*}),
\end{equation}
where $\mathbf{W}_{1}^c$ and $\mathbf{W}^p$ are initialized from the first stage, pre-trained parameters $\mathbf{W}^{e*}$ of the evaluator $\mathcal{F}^{e}$ are fixed.

The overall training loss of the second stage can be formulated as:
\begin{equation}
\label{eq:10}
    \mathcal{L} = \mathcal{L}_{e}(y, \hat{y}) + \alpha\mathcal{L}_{w}(p^a) + \beta \mathcal{L}_{a}(p^a),
\end{equation}
where we adopt Eq.~\ref{eq:5} as the evaluation loss $\mathcal{L}_{e}$. $\mathcal{L}_{w}$ is the weighted entropy loss, $\mathcal{L}_{a}$ is the area loss. $\alpha$ and $\beta$ are the hyper-parameters. In addition, in order to get a more complete activation map of the object, attentive erasing is designed and applied in model training. 
\begin{figure}[t]
\begin{center}
   \includegraphics[width=\linewidth]{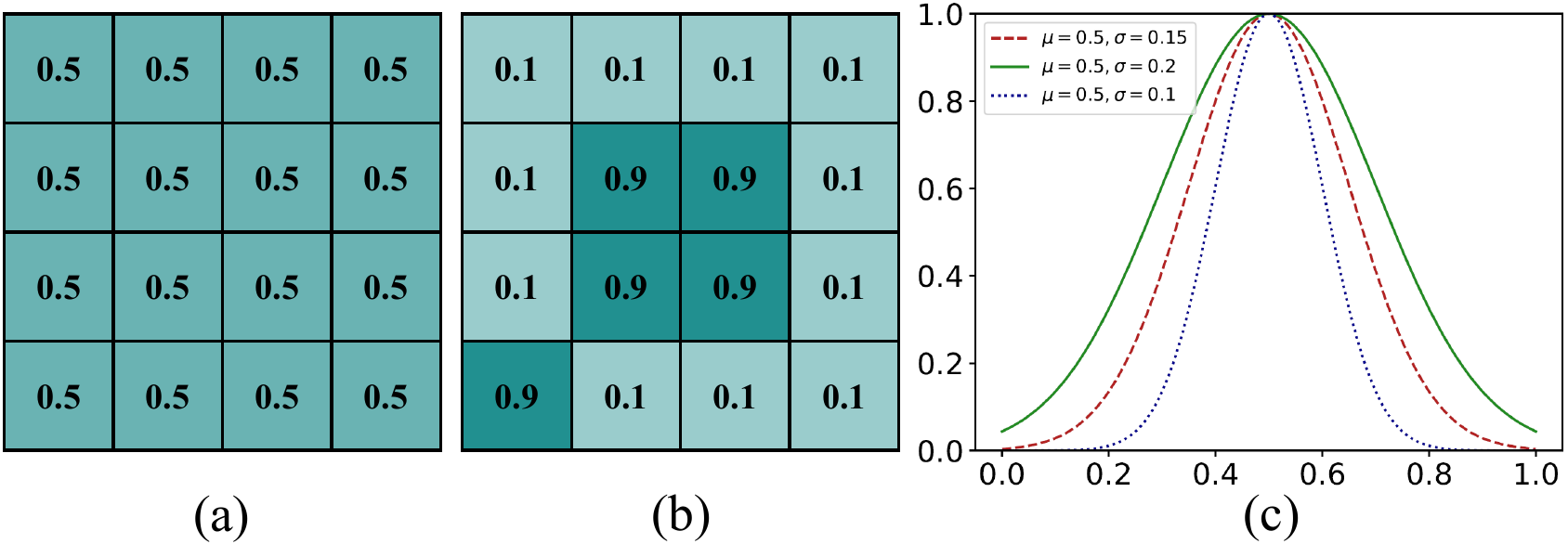}
\end{center}
    \vspace{-21pt}
    \captionsetup{font=small}
   \caption{Examples of activation maps learned without (a) and with (b) entropy loss. (c) illustrates the weighted function with different parameter settings, \ie $\mu,\sigma$. }
\label{fig:entropy_grid}
    \vspace{-11pt}
\end{figure}
\subsubsection{Weighted Entropy Loss}
Entropy, a measure of uncertainty of random variable $X$, is defined as:
\begin{equation}
H(X) = -\sum_{x \in X}P(x)\log P(x),
\end{equation}
where $P(x)$ is the occurrence probability of event $x$. Minimizing the entropy $H(X)$ can reduce the uncertainty of $X$. Based on this observation, we propose the entropy loss to reduce the uncertainty of activation map $p^a$. The $(i,j)^{th}$ pixel in the activation map $p^a$ is represented as a random variable $X_{i,j}$, which takes the values 1 (\ie foreground) and 0 (\ie background). Hence, $P(X_{i,j}=1)=p^a_{i,j}, P(X_{i,j}=0)=1-p^a_{i,j}$. Specifically, the entropy of a single pixel can be formulated as:
\begin{equation}
\label{eq:12}
     H(X_{i,j}) = - p^a_{i,j}\log (p^a_{i,j}) - (1-p^a_{i,j})\log (1-p^a_{i,j}),
\end{equation}
where $p^a_{i,j}$ indicates the $(i,j)^{th}$ element of $p^a$. Then the entropy loss of $p^a$ is defined as the mean of entropy of $X_{i,j}$:
\begin{equation}
\label{eq:13}
     \mathcal{L}_{h}(p^a) = \frac{1}{hw}\sum_{i=1}^{h}\sum_{j=1}^{w}H(X_{i,j}).
\end{equation}
While it's difficult to predict foreground/background for pixels with activation around 0.5, they are the main contributions of uncertainty. This motivates us to extend Eq.~\ref{eq:13} to our weighted entropy loss, in which the Gaussian distribution shown in Fig.~\ref{fig:entropy_grid}(c), is used to assign adaptive weight to pixels with bigger uncertainty. Specifically, the proposed weighted entropy loss is:
\begin{equation}
    \label{eq:14}
     \mathcal{L}_{w}(p^a) = \frac{1}{hw}\sum_{i=1}^{w}\sum_{j=1}^{w} \gamma_{i,j}\cdot H(X_{i,j}),
\end{equation}
where $\gamma_{i,j}$ is defined as:
\begin{equation}
    \gamma_{i,j} = e^{{-\frac{(p^a_{i,j}-\mu)^2}{2{\sigma}^2}}},
\end{equation}
where $\sigma$ and $\mu$ represent the variance and mean of the Gaussian Distribution, respectively. In our setting, $\sigma=0.1$ and $\mu=0.5$.

With the co-supervision of evaluation loss and weighted entropy loss, the generator tends to excite the foreground pixels and suppress the background pixels, which generates exact and well-separated activation maps (Fig.~\ref{fig:cm_cmp}(c)). Fig.~\ref{fig:entropy_grid} shows the examples of activation map learned without (a) and with (b) the entropy loss, one can observe that the number of uncertain pixels (0.5) has been substantially reduced by the proposed entropy loss.
\begin{figure}[t]
	\begin{center}
		\includegraphics[width=\linewidth]{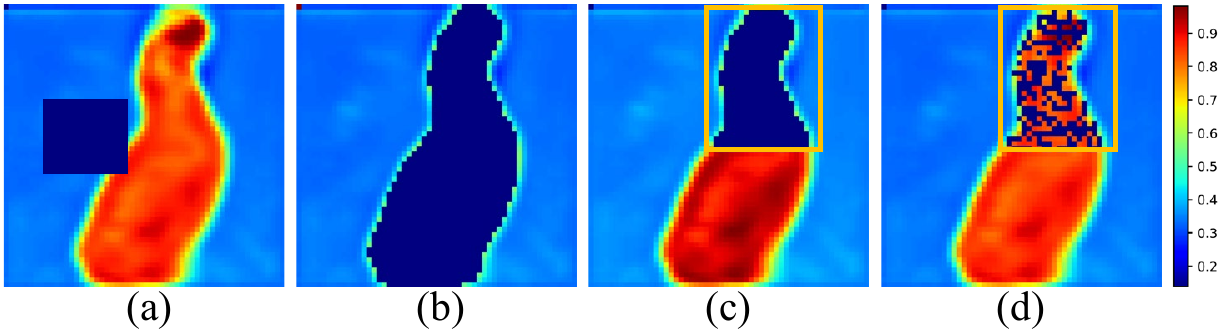}
	\end{center}
	\vspace{-21pt}
	\captionsetup{font=small}
	\caption{Various erasing strategies. (a) Random erasing. (b) Erasing the regions with activation higher than a threshold. (c) Erasing the regions within a restricted rectangle. (d) The proposed attentive erasing.}
	\label{fig:dropout}
	\vspace{-11pt}
\end{figure}
		
		
		
		
		


    
		
		
		
		
		
		
		
		
		
		
		
		
		
		
\subsubsection{Attentive Erasing}
With image-level labels, classification network mainly emphasizes the most discriminative object regions. However, the completed region of object is needed for accurate localization. A number of erasing strategies have been proposed to encourage the network to explore less discriminative regions of the object, during the generation of activation map. Fig.~\ref{fig:dropout} compares a number of available erasing strategies. As shown in the figure, random region erasing could involve the background pixels (Fig.~\ref{fig:dropout}(a)) and erasing the region with activation higher than a threshold sometimes causes the removal of whole object (Fig.~\ref{fig:dropout}(b)).

To further improve the completeness of our low-level feature based activation map, we design a so called attentive erasing in this paper, to randomly erase regions at pixel level. Firstly, we choose the coordinate of peak response in $p^a$ as the center to produce a rectangle with a random height and width. As shown in Fig.~\ref{fig:dropout}(c), the rectangle aims to restrict the erasing region, such that only pixels with value higher than a threshold and fall into the rectangle are considered as the candidates for erasing. The candidate pixels are then dropped with a probability of 0.5 (Fig.~\ref{fig:dropout}(d)).

Compared to other strategies in Fig.~\ref{fig:dropout}, our approach can encourage the network to explore entire object and retain part of the discriminative region at the same time.

\subsubsection{Area Loss}
Until now, there is no area constraint on the activation map $p^a$, which may lead to yield overlarge bounding box for inaccurate object localization. To solve this problem, we propose the area loss:
\begin{equation}
    \mathcal{L}_{a}(p^a) =\frac{1}{hw}\sum_{i=1}^{h}\sum_{j=1}^{w} p^a_{i,j}.
\end{equation}
This encourages the generator to reduce the excitation of irrelevant background clutters and ensures the compactness of activation maps for accurate object localization.

In the entropy-guided refinement stage, both the network architecture and components are designed to adversarially explore the object information preserved in the low-level features for accurate object localization. Fig.~\ref{fig:cm_cmp}(b) and (c) gives examples of the activation map generated from stage one and stage two. It can be found that a more accurate, well-separated, complete, and compact activation map can be generated after the refinement.

 \begin{figure}[t]
\begin{center}
   \includegraphics[width=\linewidth]{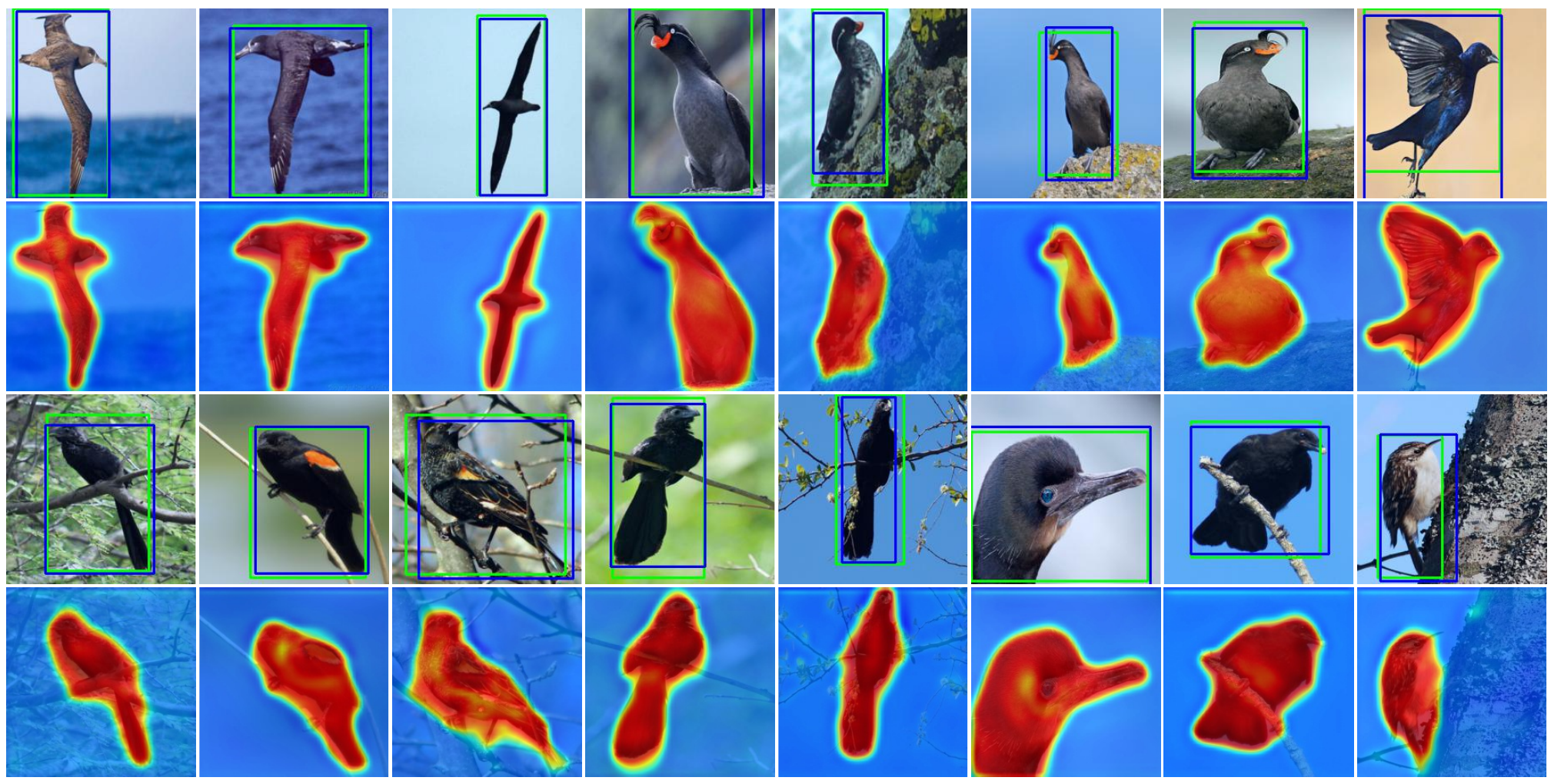}
\end{center}
\vspace{-21pt}
\captionsetup{font=small}
   \caption{Visualization of the localization results and refined activation maps on CUB-200-2011. Ground-truth and predicted bounding boxes are highlighted in blue and green, respectively.}
\label{fig:bbox_cmap}
\vspace{-11pt}
\end{figure}

 \begin{figure}[t]
\begin{center}
   \includegraphics[width=\linewidth]{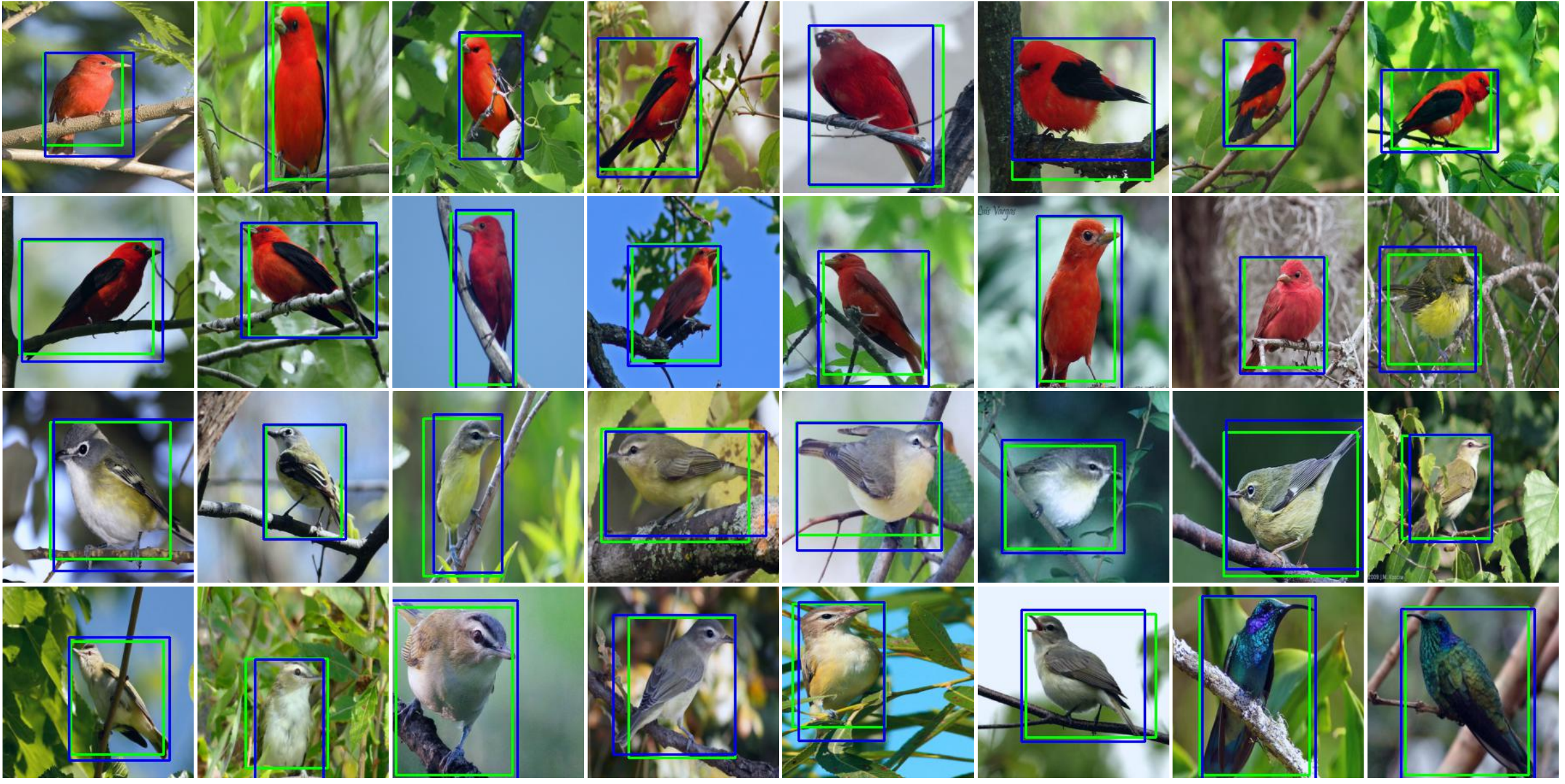}
\end{center}
\vspace{-21pt}
\captionsetup{font=small}
   \caption{Visualization of the localization results on CUB-200-2011. Ground-truth and predicted bounding boxes are highlighted in blue and green, respectively. }
\label{fig:bbox}
\vspace{-11pt}
\end{figure}

\begin{table}[t]
	\centering
    \resizebox{\linewidth}{!}{
	\begin{tabular}{lrrrrr}  
		\toprule
		&\multicolumn{2}{c}{ClsErr} &\multicolumn{2}{c}{LocErr}\\
		\cmidrule(lr){2-3} \cmidrule(lr){4-5}
		Methods compared & Top1 & Top5 & Top1 & Top5 & CorLoc\\
		\midrule
		CAM-GoogLeNet~\cite{zhou2016learning}            & 26.2 & 8.5 & 58.94 & 49.34& 55.1     \\
		Friend or Foe-GoogLeNet~\cite{xu2016friend}       &-&-&-&-& 56.5      \\
		SPG-GoogLeNet~\cite{zhang2018self}            &   -  &  -  & 53.36 & 42.28 & -    \\
		GC-Net-Elli-GoogLeNet~\cite{GCNWSOJ}    & 23.2 & 6.6 & 43.46 & 31.58 & 72.6  \\
		GC-Net-Rect-GoogLeNet~\cite{GCNWSOJ}    & 23.2 & 6.6 & 41.42 & 29.00& 75.3      \\
		DA-Net-Inception-V3~\cite{xue2019danet}           & 28.8 & 9.4 & 50.55 & 39.54& 67.0    \\
		\midrule
		CAM-VGG16~\cite{zhou2016learning}    & 23.4 & 7.5 & 55.85 & 47.84 & 56.0    \\
		ACoL-VGG16 ~\cite{zhang2018adversarial}                 & 28.1 &  -  & 54.08 & 43.49 &54.1     \\
		TSC-VGG16~\cite{he2017weakly}   & - & - & -& - & 65.5      \\
		ADL-VGG16~\cite{choe2019attention}   & 34.7 & - & 47.64& - & -      \\
		DA-Net-VGG16~\cite{xue2019danet}                 & 24.6 & 7.7 & 47.48 & 38.04 &67.7     \\
		RCAM-VGG16~\cite{bae2020rethinking}               & 29.9 & - & 42.63 & - & 78.6 \\      
		I$^2$C-VGG16~\cite{i2c} &- & - & 44.01 & 31.66 & - \\
		GC-Net-Elli-VGG16~\cite{GCNWSOJ}          & 23.2 & 7.7 & 41.15 & 30.10 & 74.9   \\
		GC-Net-Rect-VGG16~\cite{GCNWSOJ}          & 23.2 & 7.7 & 36.76 & 24.46 &81.1    \\
		\midrule
		\textbf{Ours(ORNet)-VGG16}           & \textbf{23.0} & \textbf{7.0} & \textbf{32.26} & \textbf{19.23} &\textbf{86.2}    \\
		
		\bottomrule
	\end{tabular}}
	\vspace{-11pt}
	\captionsetup{font=small}
	\caption{Comparison of the performance between the proposed method and the state-of-the-art on CUB-200-2011 test set. Here ‘ClsErr’ represents the classification error.}
	\label{tab:cub200_compar}
	\vspace{-11pt}
\end{table}

\begin{table}[t]
	\centering
	\resizebox{\linewidth}{!}{
	\setlength\tabcolsep{10pt}
	\begin{tabular}{lrrrrr}  
		\toprule
		&\multicolumn{2}{c}{ClsErr} &\multicolumn{2}{c}{LocErr}\\
		\cmidrule(lr){2-3} \cmidrule(lr){4-5}
		Methods compared  & Top1 & Top5 & Top1 & Top5 & CorLoc\\
		\midrule
		
		Backprop-GoogLeNet~\cite{simonyan2013deep}         & - & - & 61.31 & 50.55    & -  \\
		GMP-GoogLeNet~\cite{zhou2016learning}            & 35.6 & 13.9 & 57.78 & 45.26   &   -\\
		CAM-InceptionV3~\cite{zhou2016learning}              & - & - & 53.71 & 41.81   & 62.68  \\
		HaS-32-GoogLeNet~\cite{singh2017hide}          & - & - & 54.53 & -     \\
		SPG-InceptionV3~\cite{zhang2018self}             & - & - & 51.40 & 40.00   & 64.69  \\
		ACol-GoogLeNet~\cite{zhang2018adversarial}            & 29.0 & 11.8 & 53.28 & 42.58   & -  \\
		DA-Net-InceptionV3~\cite{xue2019danet}           & 27.5 & 8.6 & 52.47 & 41.72   &  - \\
		GC-Net-Rect-InceptionV3~\cite{GCNWSOJ}  & \textbf{22.6} & \textbf{6.4} & 50.94 & 41.91  & -  \\	
		
		\cmidrule(lr){1-6}
		Backprop-VGG16~\cite{simonyan2013deep}       & - & - & 61.12 & 51.46  &  -  \\
		CAM-VGG16~\cite{zhou2016learning}           & 33.4 & 12.2 & 57.20 & 45.14  &-    \\
		ACol-VGG16~\cite{zhang2018adversarial}          & 32.5 & 12.0 & 54.17 & 40.57    &  62.96\\
		ADL-VGG16~\cite{choe2019attention}   & 30.5 & - & 55.08 & - & -      \\
		
	    RCAM-VGG16~\cite{bae2020rethinking}                & 34.8 & - & 43.91 & - & 61.48 \\  
	    I$^2$C-VGG16~\cite{i2c} &30.6 & 10.7& 52.59 & 41.49 & 63.90 \\
		PSOL-VGG16-Sep~\cite{zhang2020rethinking} &- & -& 49.11 &39.10 &64.03 \\
		\midrule
		\textbf{Ours(ORNet)-VGG16}   &28.4 & 9.6 & \textbf{47.95} & \textbf{36.06} & \textbf{68.27}  \\
		\bottomrule
	\end{tabular}}
	\vspace{-11pt}
	\captionsetup{font=small}
	\caption{Comparison of the performance between the proposed method and the state-of-the-art on the ImageNet-1K validation set. Here ‘ClsErr’ represents the classification error. }
	\label{tab:ImageNet_compar}
	\vspace{-8pt}
\end{table}
\section{Experiments}
We evaluate the proposed method on CUB-200-2011~\cite{wah2011caltech} and ImageNet-1K~\cite{russakovsky2015imagenet} datasets. Extensive experiments show that our approach consistently achieves significant improvements on these benchmarks. Besides, the quantitative analysis is also conducted to verify the effectiveness of each component. We also apply our localization model to person re-identification datasets like Market-1501~\cite{zheng2015scalable}, Duke-MTMC~\cite{Ristani2016PerformanceMA} and MSMT17~\cite{wei2018person} and fine-grained classification datasets like Standford Dog~\cite{KhoslaYaoJayadevaprakashFeiFei_FGVC2011} and FGVC-Aircraft~\cite{maji2013fine}, and show the visual results in supplementary materials. The results clearly justify that our model can achieve robust localization results across a large number of different objects.
 \begin{figure}[t]
	\begin{center}
		\includegraphics[width=\linewidth]{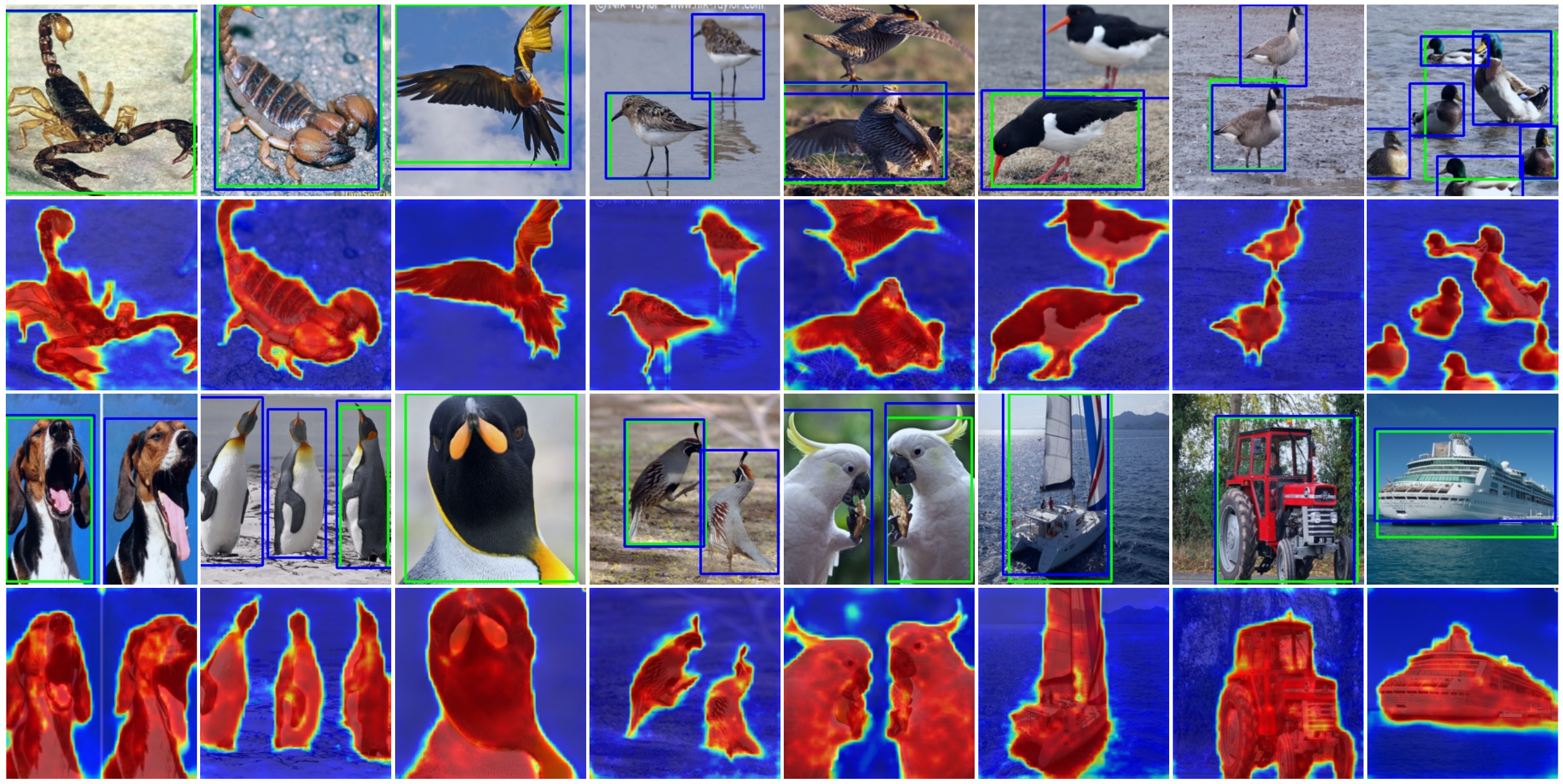}
	\end{center}
	\vspace{-21pt}
	\captionsetup{font=small}
	\caption{Visualization of localization results and refined activation maps on ImageNet-1K dataset. Ground-truth and predicted bounding boxes are highlighted in blue and green, respectively.}
	\label{fig:bbox_camp_2}
	\vspace{-11pt}
\end{figure}
\subsection{Experimental Setup}
\textbf{Datasets.} CUB-200-2011~\cite{wah2011caltech} is a fine-grained classification dataset with 200 species of birds. It consists of 11788 images, which is divided into 5994 images for training and 5794 images with bounding box annotations for testing. ImageNet-1K~\cite{russakovsky2015imagenet} contains images from 1000 object categories, which is split into 1.3 million images for training and 50000 images for testing. We only use the bounding boxes annotations of testing images for evaluation.

\textbf{Implementation details.}
 All the training images are first resized to $256\times 256$ and then augmented by random cropping to $224\times 224$. We adopt Adam as the default optimizer, with a weight decay 0.0. We adopt the cosine annealing policy~\cite{IlyaSGDR} to schedule the learning rate. Moreover, for CUB-200-2011 dataset, the mini-batch size is 16. The initial learning rate is 0.001 for the first stage and 0.0001 for the second stage. The number of training epoch is 100, including 70 epochs for stage one and 30 epochs for stage two. For ImageNet-1K dataset, we set mini-batch size to 256. The initial learning rate is 0.001 for stage one and 0.0001 for stage two, and the number of training epoch is 8, \ie 5 epochs for stage one and 3 epochs for stage two. Our model is implemented in PyTorch~\cite{paszke2019pytorch} and trained on NVIDIA Tesla P100 GPU with a 16GB memory.
 
\textbf{Evaluation metrics.} Following~\cite{deselaers2012weakly, russakovsky2015imagenet}, we adopt the localization error (LocErr), correct localization (CorLoc), Top1 accuracy, and Top5 accuracy as the metrics for evaluating the performances of the proposed methods. The LocErr is calculated based on the Top1 accuracy and location accuracy, a prediction is correct only if both localization($\ie$ IoU $\geq$ 0.5) and classification are correct. For CorLoc metric, the prediction is correct as long as the localization is correct($\ie$ IoU $\geq$ 0.5).

\begin{table}[t]
	\centering
	\scriptsize
	\resizebox{\linewidth}{!}{
	\setlength\tabcolsep{10pt}
	\begin{tabular}{lrrr}  
		\toprule
		&\multicolumn{2}{c}{LocErr}\\
		\cmidrule(lr){2-3}
		Components & Top1 & Top5 & CorLoc \\
		\midrule
		Stage one            & - & - & 64.52 \\
		\midrule
		${\cal {L}}_e$            & 44.36 & 33.02& 71.35 \\
	${\cal {L}}_e$+$\alpha {\cal {L}}_w$       & 36.05 & 23.40 & 81.68 \\
	${\cal {L}}_e$+${\beta \cal {L}}_a$       & 44.01 & 32.40 & 72.11\\
	${\cal {L}}_e$+${\cal {AE}}$       & 45.75 & 35.02 & 69.14 \\
	${\cal {L}}_e$+$\alpha{\cal {L}}_w$+$\beta{\cal {L}}_a$   & 37.31 & 24.92 & 80.10 \\
	${\cal {L}}_e$+$\alpha{\cal {L}}_w$+${\cal {AE}}$   & 33.21 & 20.21 & 85.09 \\
	${\cal {L}}_e$+${\beta\cal {L}}_a$+${\cal {AE}}$   & 39.99 & 27.91 & 76.68 \\
	${\cal {L}}_e$+$\alpha{\cal {L}}_w$+${\beta\cal {L}}_a$+${\cal {AE}}$   & \textbf{32.26} & \textbf{19.23} & \textbf{86.19} \\
		\bottomrule
	\end{tabular}}
	\vspace{-11pt}
	\captionsetup{font=small}
	\caption{Comparison of the object localization performance on CUB-200-2011 dataset using different components, including ${\cal {L}}_e$, ${\cal {L}}_w$, ${\cal {L}}_a$, and ${\cal {AE}}$.}
	\label{tab:loss_compare}
	\vspace{-8pt}
\end{table}
\begin{table}[t]
	\centering
	\scriptsize
	\resizebox{\linewidth}{!}{
	\setlength\tabcolsep{10pt}
	\begin{tabular}{lrrr}  
		\toprule
		&\multicolumn{2}{c}{LocErr}\\
		\cmidrule(lr){2-3}
		Components & Top1 & Top5 & CorLoc \\
		\midrule
		${\cal {L}}_e$+${\beta\cal {L}}_a$+${\cal {AE}}$ + $\alpha \mathbf{{\cal {L}}_h}$   & 42.53 & 31.34 & 72.89 \\
	    ${\cal {L}}_e$+${\beta\cal {L}}_a$+${\cal {AE}}$ + $\alpha \mathbf{{\cal {L}}_w}$  & \textbf{32.26} & \textbf{19.23} & \textbf{86.19} \\
		\bottomrule
	\end{tabular}}
	\vspace{-11pt}
	\captionsetup{font=small}
	\caption{The localization performance of entropy loss $\mathcal{L}_h$ and weighted entropy loss $\mathcal{L}_w$ on CUB-200-2011 dataset.}
	\label{tab:entropy_compare}
	\vspace{-8pt}
\end{table}

\subsection{Comparison to State-of-the-Arts}
\textbf{Visual Comparison.} As shown in Fig.~\ref{fig:cam_adl_ours},  CAM~\cite{zhou2016learning} mostly focuses on the most discriminative parts of objects (\eg bird head). ADL~\cite{choe2019attention} alleviates the reliance on representative features. However, it does not fully address the intrinsic defects of CAM. In contrast, in our two-stage setting, the activation maps are exact, well-separated, and compact. Fig.~\ref{fig:bbox_cmap} and Fig.~\ref{fig:bbox_camp_2} are the visual results of our method on CUB-200-2011 and ImageNet-1K, respectively, which also prove the advantages of our approach. Besides, Fig.~\ref{fig:bbox} reveals that our algorithm is also robust to locate objects even in a noisy environment. More visual results on other datasets are provided in supplementary materials.

\begin{figure*}
	\centering
	\subfigure[Analyses of hyper-parameters $\alpha$ and $\beta$.]{
		\begin{minipage}[t]{0.33\linewidth}
			\centering
		\includegraphics[width=0.9\linewidth]{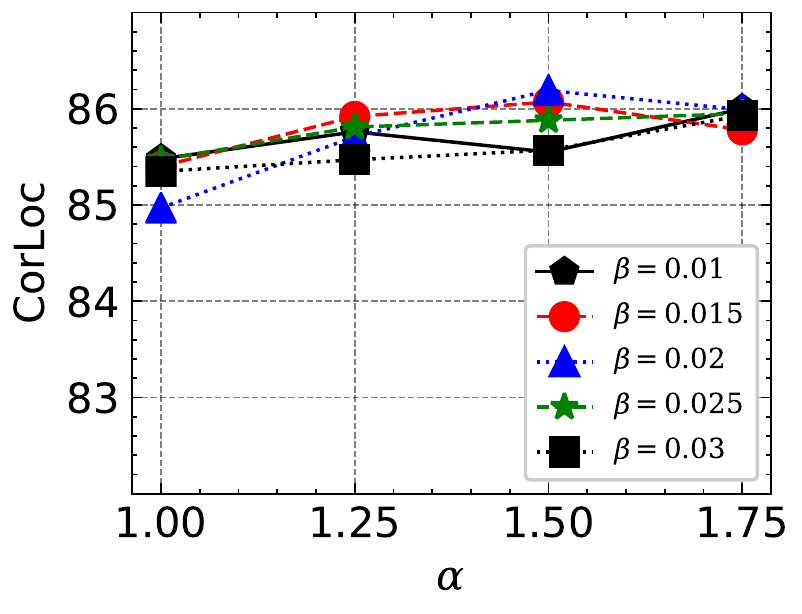}
		\end{minipage}%
	}
	\subfigure[Thresholding analyses on CUB-200-2011.]{
		\begin{minipage}[t]{0.33\linewidth}
			\centering
			\includegraphics[width=0.9\linewidth]{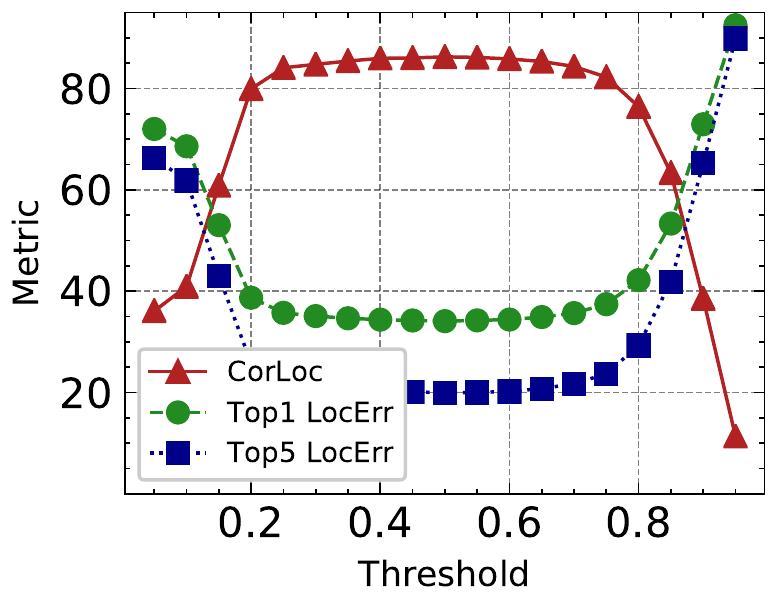}
		\end{minipage}%
	}
	\subfigure[Thresholding analyses on ImageNet-1K.]{
		\begin{minipage}[t]{0.33\linewidth}
			\centering
			\includegraphics[width=0.9\linewidth]{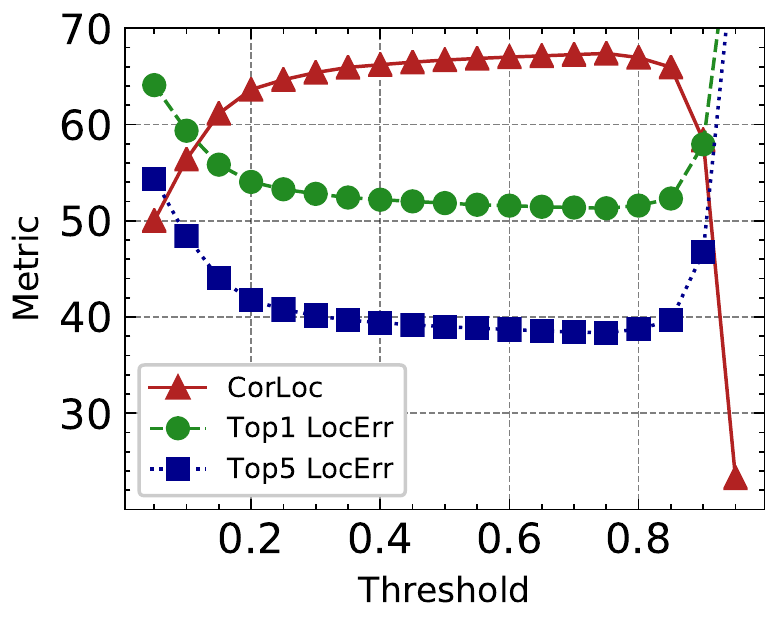}
		\end{minipage}%
	}%
	\vspace{-10pt}
	\captionsetup{font=small}
	\caption{Sensitive analyses of parameters $\alpha$ and $\beta$ and the thresholding procedure. }
	\label{fig:sensitivity_analysis}
	\vspace{-11pt}
\end{figure*}

\textbf{Quantitative Comparison}. Table~\ref{tab:cub200_compar} shows the comparison results with state-of-the-art methods on CUB-200-2011~\cite{wah2011caltech} dataset. As shown in the table, our method has the best localization performance in terms of all evaluation metrics. Specifically, our method achieves significant improvements of $5.1\%$ on CorLoc and $4.5\%$ on Top1 Loc Err over state-of-the-art GCNet~\cite{GCNWSOJ}. Compared with CAM-VGG, our method has a large improvement of $30.2\%$ and $23.59\%$ on CorLoc and Top1 LocErr, respectively.  

Table~\ref{tab:ImageNet_compar} provides the results on ImageNet-1K~\cite{russakovsky2015imagenet} dataset. Our method achieves $68.27\%$ CorLoc, outperforming all the considered state-of-the-art methods (\eg. PSOL~\cite{zhang2020rethinking}, I$^2$C~\cite{i2c}) by a large margin($\geq$ $4.16\%$). When VGG 16 is considered, our model also achieves the best classification accuracy than other approaches like CAM and ADL. As both GC-Net and our model directly take the publically available backbone trained by ImageNet, \ie Inception-V3 or VGG-16, as the classifier, the ClsErr is independent with the generator of activation map. As a result, the classification accuracy of our VGG-16 trained by ImageNet is lower than that of Inception V3 used in GC-Net. However, our model still achieves much lower Top1 LocErr ($47.95\%$) and Top5 LocErr ($36.06\%$) than those of DA-Net-Inception V3 ($52.47\%$ and $41.72\%$) and GC-Net-Inception V3 ($50.94\%$ and $41.91\%$). Top1 and Top5 LocErr consider both location and classification accuracy, \ie. a prediction is correct only if both localization and classification are correct. While fewer number of images are correctly classified, the results of LocErr suggest that our model achieves significantly higher location accuracy than DA-Net and GC-Net. 


\subsection{Ablation Study}
We now perform ablation studies using CUB-200-2011~\cite{wah2011caltech} dataset to evaluate the effectiveness of different loss functions and attentive erasing.

We first investigate the effectiveness of different combinations of the proposed components. Table~\ref{tab:loss_compare} shows the quantitative contribution of each loss function and the attentive erasing (\ie the evaluation loss $\mathcal{L}_e$, the weighted entropy loss $\mathcal{L}_w$, the area loss $\mathcal{L}_a$ and the attentive erasing $\mathcal{AE}$). As shown in the table, the generator trained in the first stage only achieves $64.52\%$ CorLoc score. However, in the second stage, a substantial improvement of CorLoc, \ie $6.83\%$, is gained by integrating the evaluator $\mathcal{F}^e$ supervised by $\mathcal{L}_e$. This verifies the necessity and effectiveness of $\mathcal{F}^e$. Besides,  $\mathcal{L}_w$ solely can also significantly promote the performance of our model, \ie $10.33\%$ improvement of CorLoc, $8.31\%$ and $9.62\%$ decreases in Top1 LocErr and Top5 LocErr. This reveals that well-separated confidence maps essentially reduce ambiguity between foreground and background regions, which benefits the thresholding procedure for accurate localization. As expected, integrating either $\mathcal{L}_a$ or $\mathcal{AE}$ fails to improve the performance, as these two components are complementary to each other. $\mathcal{L}_e$+$\mathcal{L}_a$+$\mathcal{AE}$ could significantly boost CorLoc (from $71.35\%$ to $76.68\%$) and decrease LocErr score. Moreover, $\mathcal{L}_e$+$\mathcal{L}_w$+$\mathcal{L}_a$, $\mathcal{L}_e$+$\mathcal{L}_w$+$\mathcal{AE}$, or $\mathcal{L}_a$+$\mathcal{L}_a$+$\mathcal{AE}$ also improves the performance due to the effectiveness of weighted entropy loss $\mathcal{L}_w$ or combination of $\mathcal{L}_a$ and $\mathcal{AE}$. The combination of three loss functions $\mathcal{L}_e, \mathcal{L}_w, \mathcal{L}_a$ and the attentive erasing $\mathcal{AE}$, further increase CorLoc from $71.35\%$ to $86.19\%$, and decrease Top 1 LocErr and Top 5 LocErr from $44.36\%$ to $32.26\%$ and from $33.02\%$ to $19.23\%$, respectively.

Table~\ref{tab:entropy_compare} lists the performance of activation map refined using weighted entropy loss $\mathcal{L}_w$ (Eq.~\ref{eq:14}) and mean entropy loss $\mathcal{L}_h$ (Eq.~\ref{eq:13}). As shown in the table, $\mathcal{L}_w$ increase CorLoc from $72.89\%$ to $86.19\%$, and decrease Top 1 LocErr and Top 5 LocErr from $42.53\%$ to $32.26\%$ and from $31.34\%$ to $19.23\%$, respectively. 

\subsection{Sensitivity Analysis}
There are two hyper-parameters $\alpha$ and $\beta$ in Eq.~\ref{eq:10}. The sensitivity analyses of these two parameters are performed on CUB-200-2011 test set and the results are presented in Fig.~\ref{fig:sensitivity_analysis}(a). As seen, stable CorLoc performance is obtained by a wide range setting of $\alpha$ and $\beta$. This suggests the lower sensitivity to hyper-parameters setting and good performance stability of our method. In experiments, the default value of $\alpha$ and $\beta$ are 1.5 and 0.02.

Fig.~\ref{fig:sensitivity_analysis}(b) and (c) show the sensitivity analysis of thresholding on CUB-200-2011 test set and ImageNet-1K validation set. As seen, the performances of our method are stable when the threshold varies from 0.2 to 0.8, which proves that the weighted entropy loss $\mathcal{L}_{w}$ can substantially reduce the number of uncertain pixels (0.5) and improve the robustness of our method to the setting of threshold.
\section{Conclusion and Discussion}
This article proposes a two-stage learning framework to explore low-level features based activation maps for weakly supervised object localization. The first stage uses the low-level features to yield activation map of the target object with rich contextual information. The second stage refines the activation map based on the proposed weighted entropy loss, leading to an accurate pixel-level object localization. Experiments on CUB-200-2011 and ImageNet-1K datasets validate the effectiveness of our framework. Our work suggests a promising alternative for weakly supervised semantic and instance segmentation. 
\section*{Acknowledgments}
The work is supported by the National Natural Science Foundation of China under Grant 91959108.

{\small
\bibliographystyle{ieee_fullname}
\bibliography{egbib}
}

\end{document}